\documentclass{article}
\usepackage[dvipdfmx]{graphicx}
\usepackage{arxiv}

\usepackage[utf8]{inputenc} 
\usepackage[T1]{fontenc}    
\usepackage{hyperref}       
\usepackage{url}            
\usepackage{amsmath,amsfonts}       
\usepackage{nicefrac}       
\usepackage{cleveref}       
\usepackage{doi}

\usepackage{colortbl}
\usepackage{listings}

\usepackage{parskip}

\usepackage{booktabs}
\usepackage{threeparttable}
\usepackage{tabularx}
\usepackage{multirow}

\usepackage{seqsplit}
\newcommand{\ccol}[2]{ \multicolumn{#1}{c}{#2}}

\newcommand{\myPaperTitle}{Dynamic Sentiment Analysis with Local Large Language Models using Majority Voting: \\A Study on Factors Affecting Restaurant Evaluation}
\newcommand{\myPaperShortTitle}{Dynamic Sentiment Analysis with Local Large Language Models using Majority Voting}
\title{\myPaperTitle}

\date{}

\newif\ifuniqueAffiliation

\usepackage{authblk}

\newbox{\orcid}\sbox{\orcid}{\includegraphics[scale=0.06]{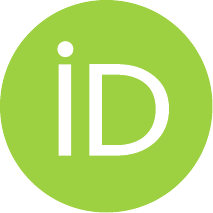}} 
\author[1,2]{%
	\href{https://orcid.org/0000-0002-4618-6272}{\usebox{\orcid}\hspace{1mm}
	Junichiro Niimi\thanks{\texttt{jniimi@meijo-u.ac.jp}}
	}}
\affil[1]{Meijo University}
\affil[2]{RIKEN AIP}

\renewcommand{\shorttitle}{\myPaperShortTitle}

\hypersetup{
pdftitle={Dynamic Sentiment Analysis with Local Large Language Models using Majority Voting: A Study on Factors Affecting Restaurant Evaluation},
pdfsubject={unsubjected},
pdfauthor={Junichiro ~Niimi},
pdfkeywords={Marketing, Natural Language Processing, Sentiment Analysis, Large Language Model, Quantization},
}

\begin{document}


\twocolumn[
	\begin{@twocolumnfalse}
		\maketitle
\vspace{-3em}
\begin{abstract}
User-generated contents (UGCs) on online platforms allow marketing researchers to understand consumer preferences for products and services.
With the advance of large language models (LLMs), some studies utilized the models for annotation and sentiment analysis. However, the relationship between the accuracy and the hyper-parameters of LLMs is yet to be thoroughly examined. In addition, the issues of variability and reproducibility of results from each trial of LLMs have rarely been considered in existing literature. Since actual human annotation uses majority voting to resolve disagreements among annotators, this study introduces a majority voting mechanism to a sentiment analysis model using local LLMs. 
By a series of three analyses of online reviews on restaurant evaluations, we demonstrate that majority voting with multiple attempts using a medium-sized model produces more robust results than using a large model with a single attempt. Furthermore, we conducted further analysis to investigate the effect of each aspect on the overall evaluation.
\end{abstract}
\vspace{0.5em}
\vspace{2em}
	\end{@twocolumnfalse}
]

\section{Introduction}\label{sec:introduction}
Nowadays, as consumers spontaneously post their opinions regarding products and services on online platforms, such as social media and review apps, user-generated contents (UGCs) are widely available in various business fields. In other words, the analysis of textual data is crucial for market research in terms of product development, service improvement,  and other activities. For example, textual data collected from online platforms and social networks have been utilized for company's decision-making, such as evaluating \cite{sentiment_marketing} and extracting \cite{extract_product_features} product features, constructing a recommendation system \cite{bert_hotel}, and assessing the effect  on purchase intention \cite{wom_to_purchase_intention}. However, even in this situation, text data are yet to be fully utilized, in contrast to the amount of accumulated data \cite{rulebased-annot2}. 

To  extract, utilize, and understand consumer preferences  from textual data, pre-processing the data through assigning labels, classifying the text, and evaluating the sentiment is essential. However, these are labor-intensive tasks for humans, whereas automated analysis using natural language processing (NLP) technology requires domain knowledge such as text mining and machine learning. Companies can choose another option to use crowdsourcing such as Amazon Mechanical Turk (MTurk), which has become widely used in recent years but is costly, particularly when dealing with large-scale data. Moreover, the quality of crowdsourced data has been a serious concern in academia \cite{mturk1,mturk2}. Thus, as the enormous amount of data has been accumulated, handling big data becomes more challenging and often impractical. 

Large language models (LLMs) have become readily available with multiple cloud services, such as ChatGPT by OpenAI (https://chatgpt.com), Gemini by Google (https://gemini.google.com), and Claude by Anthropic (https://claude.ai). These LLMs are used for a wide range of tasks, leading to the continuous development of new services and applications. In addition, several studies proposed automated annotation models using LLMs \cite{annotation_gpt3,annotation_gpt4,gptVsMTurk}. LLMs have highlighted advantages such as high processing speed, low cost, and reproducibility compared to human annotators \cite{annotation_gpt4}. Moreover, the  initial costs for computational resources and training data preparation is lower than those of machine learning. For instance, it has also been shown that ChatGPT operates at one-thirtieth the cost of MTurk \cite{gptVsMTurk}. 

However, from a practical perspective, the confidentiality of in-house data is an important issue for companies. The utilization of cloud-based LLMs in business activities may also pose significant security risks, such as information leakage or data falsification. In fact, some studies \cite{cloudSecurity1,cloudSecurity2} have reported that the protection of user data is an obstacle in introducing cloud services. Additionally, there are serious concerns about the network, compliance, and information security of the cloud environment for commercial use. In particular, companies may adopt a more conservative approach in using AI due to concerns regarding unauthorized learning of their data within the AI sector. Therefore, LLMs that operate on local computers have attracted considerable attention. However, according to the existing literature \cite{quantize_validation}, their performance varies significantly depending on hyper-parameters, such as the precision of quantization, and the relationships between these factors have not been fully explored particularly for the marketing research.

Therefore, in this study, we propose a model for sentiment analysis that can be executed on local computing resources using commercially available open-source LLMs. To analyze multiple aspects of the opinion dynamically, the model is built for aspect-based sentiment analysis (AbSA) \cite{absa_review1} with arbitrary aspects set by the authors.
Furthermore, this study is not limited to the mere extraction of information by the proposed model but demonstrates that the obtained data can be utilized for further statistical analysis. The remainder of this paper is organized as follows.
In Section 2, we review related studies to contextualize this research. In Section 3, we introduce our proposed model. An overview and the results of the analysis are presented in Section 4. Finally, in Section 5, we discuss the implications and  challenges of this study. 

\section{Related Works}
\subsection{Sentiment Analysis}
Previous studies on sentiment analysis mainly focused on electronic word-of-mouth (eWOM) and social network posts \cite{sentiment_review}. These methods can be broadly classified into four groups: rule-based models, machine learning, deep neural networks (DNNs), and LLMs. 

\subsubsection{Rule-based Models}
Several well-known models, such as valence aware dictionary for sentiment reasoning (VADER) \cite{vader}, semantic orientation calculator (SO-CAL) \cite{socal}, and TextBlob \cite{textblob}, have been proposed for rule-based sentiment analysis using lexicons. These models have been widely adopted for sentiment analysis \cite{sentiment_review,rulebased-annot3,vader1,vader2,paas}. 
For the advantages, these models do not require a training process since all the evaluations are based on the pre-determined rules and lexicons. these models have high interpretability for the obtained results while machine learning in general is considered as a black-box model \cite{vader,annotation_review1}. In addition, the duration of inference was significantly shorter than that of the other methods.

However, several challenges remain. For example, a rule-based model assumes that all words used are included in the dictionary, making it difficult to deal with unknown words. Moreover, dynamic capturing of sentiments is difficult because the applicability of negation in a sentence is assessed based on a rule \cite{socal}. As for the major problem, the models can only adjudicate polarity across the entire text; hence, they cannot measure individual perspectives as AbSA. Considering these characteristics, while rule-based models certainly have significant advantages in being easy, fast, and inexpensive to implement, they struggle to perform sentiment analysis flexibly from an individual perspective. Therefore, a more advanced approach is required.

\subsubsection{Machine Learning}
Another approach in sentiment analysis is the use of machine-learning techniques \cite{ml-annot1,ml-annot2}. Major machine learning techniques, such as k-nearest neighbor (kNN) \cite{knn}, Na\"{i}ve Bayes (NB), and linear support vector machine (linear SVM) \cite{svm}, can perform sentiment analysis of textual data as part of the classification task. To handle text using these models, first converting the sentence into word embeddings is necessary. For example, the well-known approaches are term frequency-inverse document frequency (TF-IDF) \cite{tfidf}, word2vec \cite{word2vec}, and FastText \cite{fasttext1,fasttext2}. A previous study \cite{nb_kNN} which implemented sentiment analysis on the review data of movies and hotels  compared the accuracies between kNN and NB. Another study \cite{nb_hotel} which classified consumers based on online reviews according to the extent of satisfaction with hotels adopted NB. Further study \cite{tfidf-svm} which conducted sentiment analysis on multiple datasets, including Amazon's online reviews, employed a combination of TF-IDF and linear SVMs.

These machine-learning techniques can estimate the polarity for individual dimensions in addition to that over the whole text as long as they have the correct labels. However, correct labels and training processes are necessary for this purpose. As mentioned previously, preparing a sufficient dataset for training is labor-intensive or costly. 
Furthermore,  other challenges exist in acquiring these features. For example, embedding methods, even word2vec, do not capture the dynamic meaning of words based on the relationship between several sentences. That is, the same word is treated constantly, even if it has different meanings, and it cannot handle word polysemy. In this situation, it is difficult to comprehend the ever-changing interests of consumers and respond flexibly to unknown perspectives that may emerge in the future, which is an important objective in the use of texts for planning marketing strategies.

\subsubsection{Deep Neural Networks}
With the advancement of neural networks, various DNN models have been proposed for sentiment analysis \cite{dnn_sent_review1,dnn_sent_review2}. Typically, DNN models also adopt embedding techniques to obtain distributed representation. First,  convolutional neural network (CNN) based method leverages their strength in capturing local patterns. A study \cite{fasttext_cnn} which constructed sentiment classification models on several text datasets adopted a combination of FastText and CNN. The model alternates between convolutional layers and max-pooling layers applied to the 2D feature representations, followed by fully connected layers with ReLU activation, and finally, classification with a softmax function. 
Second, the combination of attention mechanism \cite{attention} and recurrent layers, including recurrent neural network (RNN) \cite{rnn}, gated recurrent unit (GRU) \cite{gru}, and long short-term memory (LSTM) \cite{lstm}, has been explored in some studies \cite{attention_sentiment1,attention_sentiment2,attention_sentiment3} to develop sentiment analysis models that are sensitive to prediction-relevant information. 
Moreover, bidirectional encoder representations from transformers (BERT) \cite{bert} and their advanced models \cite{roberta,distilbert} have made substantial contributions to the textual analysis. They can obtain deep-contextualized word representations \cite{elmo} that dynamically change the meaning of a word based on its interaction with other words in the sentence. One study \cite{bert_googleplay} established a sentiment analysis model using BERT to predict sentiments of user reviews on online platforms. For AbSA, a study \cite{bert_hotel} simultaneously predicted multiple aspects of hotel evaluation, such as overall, service, and location ratings from user reviews on online platforms. They \cite{bert_hotel} not only proposed the AbSA model but also constructed a recommender based on predicted sentiments. That is, an easy-to-implement AbSA would lead to a deeper understanding of consumer preferences and certainly contribute to the construction of personalized services.

In addition, multimodal deep learning \cite{multimodal} has received increasing attention. It combines multiple data streams and considers their relationships to construct a robust predictor \cite{multimodal_survey1}. In multimodal sentiment analysis \cite{multimodal_sentiment_review1,multimodal_sentiment_review2}, a model is extended to handle modalities other than textual data, such as numeric values, images, and audio, and can utilize non-verbal information absent in text to construct relationships among the modalities. 
A study \cite{niimi_crossattention} which focused on user ratings for restaurants constructed a multimodal model that simultaneously integrates the textual data of review texts and the tabular data of user and restaurant information using BERT and cross-attention\footnote{The study \cite{niimi_crossattention} uses the same dataset as this study. However, because of different data extraction conditions, a direct comparison is not possible.}.

DNN models have a strong advantage in terms of their high prediction accuracy because of their multilayer structure and nonlinear modeling \cite{deeplearning}. For AbSA, the models can predict multiple dimensions simultaneously with an appropriate loss function. However, using deep learning has drawbacks of high computational costs in terms of both time and resources required for training the model. In general, large-scale computational resources using GPU and large amounts of data are required.

\subsubsection{Large Language Models}
Recently, a few studies utilized LLMs for the annotation of unstructured data and sentiment analysis \cite{llm_sentiment_review}. In sentiment analysis, as well as the general usage of LLMs, an analyst creates and passes the prompt which includes instruction and review texts to the LLMs. The sentiment values are extracted from the response. For example, a study \cite{annotation_gpt3} employed GPT-3 which is one of the model variants provided in ChatGPT to compare several annotation approaches. Another study \cite{gpt35_sentiment} uses GPT-3.5 for multiple analyses including sentiment prediction. 
In terms of comparison with human annotation, a previous study \cite{gptVsMTurk} reported that the zero-shot model \footnote{In the zero-shot model, a task can be performed with sufficient accuracy without instructing any examples of answers in the prompts. Similarly, the few-shot model can be executed with only a small number of examples.} outperformed MTurk's crowdworker by an average of 25 \% in terms of accuracy on the four annotation tasks. Another study \cite{gpt4_humancomparison} used GPT-4 to predict the sentiments of social media posts and reported that the predictions substantially matched the human rating values. Moreover, a recent study \cite{annotation_gpt4} also used GPT-4 model to annotate multiple datasets; however, the results presented in their study are limited to summary statistics, and it is unclear whether accuracy was achieved in specific tasks, making it difficult to assess its usefulness. 

To summarize, the first and most significant advantage of sentiment analysis using LLMs is their ease of use. These methods can be used in natural language and therefore require less expertise than any of the models described thus far. While research in this field is still relatively limited and many studies are in preprint stages, the existing body of work strongly supports the efficacy of LLMs in text classification tasks, including sentiment analysis, indicating their promising potential. In addition, analyses using LLMs tend to be more accurate than the traditional models, which may be because LLMs in general are pre-trained on a large-scale text corpus. The pretraining process makes them zero- or few-shots learners \cite{gpt3}. In other words, they have high potential for use in marketing research since they can provide sufficient accuracy in annotating unknown data without additional learning (i.e., fine-tuning). 

However, most of these models do not attempt multiple annotation trials, which rarely considers the variability and consistency in using LLMs. For business applications, the reproducibility and consistency of results are crucial. 
Even in the actual human annotation, tasks are generally conducted by multiple workers. If there is a disagreement among workers, solutions such as discussions and majority voting are adopted to determine the final evaluation. 
One study \cite{annotation_gpt4} that examined the change in accuracy by repeating LLM annotations for the multiple times showed that the higher the consistency across multiple annotations, the higher the final accuracy and recommended three or more trials for the annotation of one dataset. 
In other words, by incorporating the majority voting mechanism into the multiple attempts of LLMs, the performance of the task is expected to increase. Therefore, in this study, we develop LLM sentiment analysis model which generates multiple workers inside the model and each worker votes the sentiment to generate more robust results.

\subsection{Local LLMs and Quantization}
With the rise of online AI chat services, running LLMs on local devices, such as laptop computers and smartphones, has been explored \cite{mobile_llm}. As mentioned previously, LLMs are assumed to be executed in abundant environments. In other words, the major challenge is the limited computing resources. One approach for addressing this issue is to apply quantization \cite{quantize_int8,quantize_gptq,quantize_zeroquant,quantize_awq,llama-cpp}. Quantization is referred to as "to map floating-point numbers into low-bit integers" \cite{quantize3}, which discretizes the continuous values of the parameters on the LLMs to compress the model size and memory usage and accelerate its execution \cite{quantize}. 

Quantization can be divided into two approaches: quantization-aware training (QAT) and post-training quantization (PTQ). QAT designs the learning process based on the assumption of quantization. The inference accuracy tends to increase since the model can be adapted for quantization during the early stages of training. However, the quantization is required to be considered during the training process, needing a large amount of resources and expertise compared to PTQ \cite{quantize1}, which applies quantization after training. On the other hand, PTQ applies quantization after the training. In PTQ, a smaller amount of data is required for the calibration of the model, and its implementation is more effective than that of QAT \cite{quantize3}. In this study, we focused on PTQ---a widely used technique for quantization owing to its low training costs.

As previously mentioned, quantization generally limits the parameter $w \in \mathbb{R}$ to an integer. For example, $b$-bit quantization ($b \in \mathbb{N}$) uses the map function $\phi: (\mathbb{R}, \mathbb{N}) \to [0, 2^b)$ to obtain the quantized parameter $\hat{w}$ \cite{quantize}. The actual PTQ algorithm has variants such as LLM.int8() \cite{quantize_int8}, GPTQ \cite{quantize_gptq}, SpQR \cite{quantize_spqr}, AWQ \cite{quantize_awq}, and GGUF \cite{llama-cpp}. Whichever technique is used, precision of quantization, specifically the number of bits in quantization, should be carefully considered because it generally degrades the performance of the quantized model. In other words, the performance of the model and the precision of the quantization have a tradeoff relationship \cite{quantize_validation,quantize_balance}.

In summary, although methods to execute LLMs on resource-constrained devices have been explored, there is still a lack of verification of points such as the relationship between precision and accuracy. Additionally, it is difficult for industries to comprehensively evaluate an optimal model with a suitable balance between execution speed and accuracy in marketing analysis. Therefore, to understand this relationship, we construct AbSA models using pre-trained LLMs that have different numbers of parameters and are quantized with different precisions. Upon selecting the pretrained LLMs, we mainly focused on 4-bit quantization since several studies \cite{quantize1,quantize2} have shown that 4-bit quantization can perform close to the nonquantized model. Needless to say, it has also been pointed out that the performance of LLMs is highly dependent on the task \cite{quantize3}.

\section{Proposed Model}
\subsection{Pretrained Models}
In terms of pretrained models, we employ instruction-tuned models that require no additional training. In instruction tuning \cite{instruction-tuning}, the model is trained in advance using a combination of various instructions and their expected responses to adapt to a wide range of tasks. 
In this study, Llama provided by Meta \cite{llama2} is adopted, which is an open-source LLM permitted for both commercial and academic use only if several conditions are fulfilled. According to the model card \cite{llama3modelcard}, latest Llama 3 is trained with more than 15 trillion tokens from publicly available data sources. Furthermore, the fine-tuning process of the model includes more than 10,000 manually annotated data in addition to instructional data, as well as reinforcement learning from human feedback (RLHF) \cite{rlhf}. Considering these learning processes, sufficient accuracy is expected for sentiment analysis of eWOM without additional training.

We focused on three factors: model scale (i.e., number of parameters), precision for quantization, and model architecture.
First, in terms of model scale, the latest Llama 3 has two variants: 8 billion (8B) and 70 billion (70B). Second, in terms of precision, we primarily used 4-bit models with additional 3-bit and 5-bit models. Finally, the Llama 2 model was used to validate the impact of the model architecture and pretraining process. Table \ref{tab:models} lists the employed models. We compared the performance and duration of each model in practical marketing research tasks. For quantization, the GGUF format \cite{llama-cpp} is employed, which has been widely adopted within the LLM community because of its high practicality.

\begin{table}[htb] 
      \begin{center}
      \caption{Employed Pre-trained LLMs}
      \label{tab:models}
        \scalebox{0.75}{
\begin{tabular}{
wl{5cm}wc{0.8cm}wc{0.8cm}wc{0.8cm}wc{0.8cm}
}
\toprule
\multicolumn{1}{c}{Model Name} & Llama & Scale & Precision & PTQ \\
\midrule
Meta-Llama-3-70B-Instruct.Q4\_K\_M & 3 & 70B & 4-bit & GGUF \\
Meta-Llama-3-8B-Instruct.Q5\_K\_M & 3 & ~~8B & 5-bit & GGUF \\
Meta-Llama-3-8B-Instruct.Q4\_K\_M & 3 & ~~8B & 4-bit & GGUF \\
Meta-Llama-3-8B-Instruct.Q3\_K\_M & 3 & ~~8B & 3-bit & GGUF \\
Meta-Llama-2-7B-Chat.Q4\_K\_M & 2 & ~~7B & 4-bit & GGUF\\
\bottomrule
\end{tabular}
        }
        \end{center}
\end{table} 

In addition, instead of implementing fine tuning, we employ one-shot learning. We included a single annotated example in the prompt and enabled the model to generate an accurate response. The samples used for the instructions were randomly extracted from the dataset, annotated by the authors, and excluded from the test data. Since Llama 3 has a large context window of 8192 tokens, it is sufficient to wrap up the entire text, including the prompt, one-shot example, and review texts. To extract the polarity value as structured data, the model is explicitly required to output text in the JSON format. Once obtained, the text is parsed into tabular data. 

\subsection{Majority Voting Mechanism}
In the field of machine learning, ensemble learning is often employed to reduce the error rate and to generalize results. It is a combination of predictions by multiple models and has been shown to be robust to noise and outliers \cite{ensemble1}. The utility of learning has been widely confirmed (cf. review articles and books \cite{ensemble1,ensemble2,ensemble3}), particularly when models have different types of prediction errors. In other words, the proposed model is effective when the responses exhibit moderate diversity. 
In ensemble learning, majority voting is utilized particularly for multinomial classification, in which each model votes for one class, and the class with the most votes is considered as the final predictor \cite{ensemble2}. While ordinary majority voting used the average evaluation value, LLMs may occasionally produce unexpected values such as outliers or missing values. Therefore, in this study, we adopt the median value for the metric of majority voting. 

To implement this mechanism, the reproducibility parameters of the model were used. In most machine learning methods, by specifying the initial value of the random number generation (i.e., seed value), it is possible to ensure reproducibility in training and inference. In other words, by repeating annotations with different seed values, the model can generate the results obtained by multiple fictitious workers. Based on the responses of each virtual worker, we create two variables: 1) $m^{k}_w$, an indicator of whether dimension $k$ is mentioned in worker $w$, and 2) $v^{k}_w $, an indicator of how high the sentiment is if mentioned (where dimension $k = \{1,2,...,14\}$ and worker $w = \{1,2,..., 5\}$). 

First, for each worker $w$, 
\begin{align}
m^{k}_w &= 
   \begin{cases}
      1 & \text{(if~} k \text{~is mentioned)} \\
      0 & \text{(otherwise)}
   \end{cases} \\
v^{k}_w &= 
   \begin{cases}
      v & \text{(if~} k \text{~is mentioned)} \\
      0 & \text{(otherwise)}
   \end{cases}
\end{align}
are obtained (sentiment $v = \{1,2,..., 5\}$). Second, voting was conducted using the median as follows:
\begin{align}
m^{k} &= median(m^{k}_w | ~w=1, 2, 3, 4, 5) \\
v^{k} &= 
\begin{cases} 
median(v^{k}_w \mid v^{k}_w \neq 0) & \text{(if } \exists ~w \text{ s.t. } s^{k}_w \neq 0\text{)} \\
0 & \text{(otherwise)}
\end{cases} 
\end{align}
This implies that $v^{k}$, the median of the sentiment values for the dimension $k$, is calculated only among nonzero $v^k_i$. Finally, the voted evaluation $s^k$ was obtained as follows:
 \begin{align}
s^k = m^{k} v^{k}
\end{align}
Thus, this mechanism employs two-stage majority voting to assess the level of sentiment on the dimension $k$: 1) determining whether the dimension is mentioned or not, and 2) evaluating the sentiment polarity, utilizing the median for both stages.

In addition, the randomness of the response of LLMs is determined using a temperature parameter. One study \cite{annotation_gpt4} suggests 0.2 and above for temperature while another study \cite{gpt_temperature} pointed out that the result becomes highly random and not consistent with higher temperature. Therefore, we use a temperature value of 0.2 for a moderate randomness.

\section{Analysis}
\begin{figure*}[tbh]
   \centering
   \includegraphics[width=0.9\linewidth]{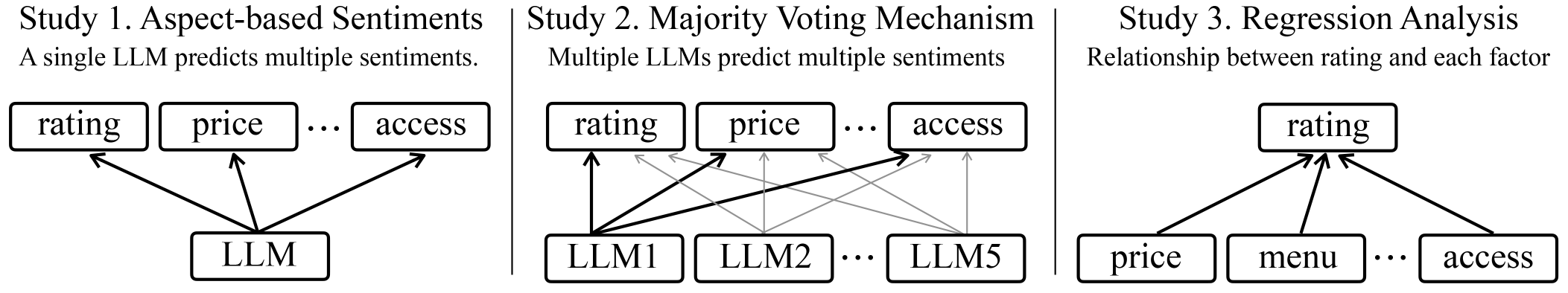}
   \caption{Overview of the Analyses}
   \label{fig:study}
\end{figure*}

To validate the proposed model, this study consists of three consecutive analyses (Study 1-3). An overview of the analysis is presented in Fig. \ref{fig:study}. First, in Study 1, we conduct a sentiment analysis for restaurant reviews posted on an online platform. In the analysis, we construct multiple models with different hyper-parameters and examine the changes depending on the settings. We do not use a majority voting mechanism here because Study 1 is an exhaustive analysis using many variants, including large-scale models.
In Study 2, we integrate the majority voting mechanism into the model chosen in Study 1. We iterate the inferences multiple times on a single task and validate the consistency among the trials and the change in accuracy through voting.
Finally, in Study 3, two linear regression models are established for further analysis using the obtained aspect-based sentiments. The two models predict the actual and predicted ratings for the restaurant using individual aspects of the evaluation, such as the price of the restaurant and taste of the dishes. We compare the estimated parameters of the models and demonstrate that the annotation results of the proposed model do not differ from the ground truth.

In Studies 1 and 2, the performance of each model was evaluated using three metrics: concordance rate (Acc.), Pearson's correlation coefficients (Corr.), and the root mean square error (RMSE). In general, classification tasks are assessed using metrics such as precision, recall, and F1 score, which assess how well the predictions align with the actual labels. However, because our target variable is ordinal, the magnitude of the prediction errors is important. If a prediction is incorrect, the extent of the error–-whether it is a large deviation or a neighboring value–-matters. Therefore, we evaluate the extent of the discrepancy between the predicted and actual values.

\subsection{Data Overview}
In this study, we use the Yelp Open Dataset \cite{yelp}, an open dataset publicly available for academic use. Yelp is an online platform on which users post evaluations and reviews about various facilities, including restaurants, stores, and public institutions. Some studies \cite{niimi_crossattention,yelp0,yelp1,yelp2} have utilized it to analyze user reviews. The data contain the ratings $nStars_{ij}$ and review texts $review_{ij}$ that user $i$ posts on facility $j$ (where $i = 1, 2, \dots, n$; $j = 1,2, \dots, m$). One post was randomly extracted from each user. If a user posted a review for the same restaurant multiple times, only the most recent review was considered for sampling. Therefore, the sample size matches the number of target users extracted.

In addition, each establishment has category tags, such as Restaurant, Coffee \& Tea, and Bar; thus, we can extract the target instances by choosing the tags. In this study, only restaurants holding a physical store in a fixed address were included in the analysis, and therefore we extracted only those tagged with Restaurant and excluded those of Fast Food, Food Trucks, Nightlife, and Bar. We sample 1000 instances from the data, and Table \ref{tab:tokens} lists the summary statistics. The numbers of characters and tokens are counted using the Tiktoken tokenizer \cite{tiktoken} which is also used in Llama 3.

\begin{table}[htb] 
      \begin{center}
      \caption{Summary Statistics of the Data}
      \label{tab:tokens}
        \scalebox{0.95}{
\begin{threeparttable}
\begin{tabular}{llcccr}
\toprule
& & \ccol{1}{Mean} & \ccol{1}{Std} & \ccol{1}{Min.} & \ccol{1}{Max.} \\
\midrule
\multicolumn{2}{l}{\bf Textual Data}\\
& \#Characters & 392.062 & 302.190 & 61 & 2425 \\
& \#Tokens & ~~88.164 & ~~67.973 & 13 & 570 \\
\midrule
\multicolumn{2}{l}{\bf Evaluation Data}\\
& \#Stars &  ~~~~3.933 & ~~~~1.371 & \ccol{1}{1} & \ccol{1}{5} \\
\bottomrule
\end{tabular}
\begin{tablenotes}[para,flushleft,online,normal] 
\end{tablenotes}
\end{threeparttable}
        }
        \end{center}
\end{table} 

\subsection{Study 1: Effects of Model Scale and Precision on the Performance}
First, we predict the overall rating (i.e., the number of stars given) of restaurants in Study 1. By comparing the accuracy and processing speed of predictions among multiple pre-training and reference models using machine learning methods, we explore the best-balanced model that optimizes both prediction accuracy and processing speed. In the analysis, annotation was performed using a 5-point Likert scale, in accordance with the fact that the overall rating by actual users ranged from one to five stars. 
In addition to overall sentiment, 14 dimensions were simultaneously predicted for Study 2 (cf. Table \ref{tab:dims}, in Study 2).

\begin{table}[htb] 
      \begin{center}
      \caption{Aspects of the Sentiment}
      \label{tab:dims}
        \scalebox{1}{
\begin{threeparttable}
\begin{tabular}{
wl{1.8cm}wl{5cm}
}
\toprule
\multicolumn{1}{c}{Dimensions} & \multicolumn{1}{c}{Explanations}\\
\midrule
$overall$ & overall rating on the restaurant\\
$price$ & price of the restaurant \\
$menu$ & variety of menu \\
$dishes$ & taste of dishes \\
$dessert$ & taste of desserts \\
$cleanliness$ & cleanliness of the restaurant \\
$atmosphere$& atmosphere of the restaurant\\
$congestion$ & congestion of the restaurant \\
$noise$ & noise in the restaurant \\
$attitude$ & attitude of other customers \\
$enjoyment$ & other entertainment service, such as \\
                      & live music, DJs, and cafe seminar \\
$child$ & child-friendliness \\
$couple$ & suitability for couples \\
$access$ & ease of access \\
\bottomrule
\end{tabular}
\begin{tablenotes}[para,flushleft,online,normal] 
\end{tablenotes}
\end{threeparttable}
        }
        \end{center}
\end{table} 

Eight reference models were established to evaluate the model performance. 
First, we construct three DNN models: feed-forward neural network (FFNN), bidirectional LSTM (Bi-LSTM), and convolutional neural network (CNN). 
The FFNN model uses BERT (pre-trained model: bert-base-uncased \cite{bert}) for text vectorization, which has the 712-dimensional pooled output of the [CLS] token from BERT and is passed through three fully connected layers and classified using a softmax function. 
The Bi-LSTM model is constructed in accordance with the previous study \cite{lstm_att}. We used word2vec for the vectorization and combined Bi-LSTM and multi-head attention, which addresses the long-term dependencies of the context and add weighted importance to the relevant information. \footnote{The actual previous study \cite{lstm_att} grouped the target variable into positive, neutral, and negative.}
The CNN model is established with the previous study \cite{fasttext_cnn} which used FastText for obtaining word embedding. The model alternates between convolutional layers and max-pooling layers applied to the 2D feature representations, followed by fully connected layers with ReLU activation, and finally, classification with a softmax function. 
Second, we used three rule-based methods, including VADER, SO-CAL, and TextBlob. These methods are based on pre-determined dictionaries and rules, so there are no adjustable parameters. Our predictions are generated directly from these polarity scores without any additional training\footnote{The obtained continuous values were adjusted to a range in $[1, 5]$.}. 
Third, for machine learning methods, we use Linear SVM. Since machine learning methods cannot directly handle the textual modality, we vectorize the text using TF-IDF in accordance with the previous study \cite{tfidf-svm}. To create sufficiently sized features for prediction with TF-IDF, we set the dimensionality to 4000. 
Finally, in addition to these well-known models, a benchmark at chance level is created. Instead of generating completely random predictions, we calculated the proportion of sentiment labels from the training data and used it as weights to generate predictions for the test data. 
Regarding the additional dataset for training and validation, we randomly sampled 5000 observations for training and 1000 for validation set, ensuring no duplication among the datasets. 

The results are shown in Table \ref{tab:result1}. The proposed models (Model 1--4) except Llama-2 (Model 5) outperformed the reference models. Notably, Llama 3 with 70B parameters (Model 1) demonstrated the best performance across all metrics. In terms of the number of parameters, the model with 70B in 4-bit (Model 1) is superior to that with 8B in 4-bit (Model 3), indicating that a larger-scale model achieves better performance within the same precision of 4-bit. In other words, the scale of the model generally contributes on the prediction accuracy as well as previous studies have shown \cite{openai-scaling-laws,llm-params1}.
Second, in terms of precision in quantization for 8B models (Models 2--4), in contrast to expectations, the accuracy improved with lower precision. Third, as all Llama 3 models (Models 1--4) outperformed the Llama 2 model (Model 5), it can be concluded that the improved model structure and pre-training process contributed to the predictions. 

Regarding the reference models, performance was in the following order: DNN models (Models 6--8), rule-based models (Models 9--11), machine learning models (Model 12), and chance level. Although the model using DNN achieved high accuracy as expected, it is noteworthy that most of them did not surpass that of Llama 2. Additionally, among the DNN models, the highest accuracy was shown when BERT was used for acquiring word embeddings (Model 6). Although Model 7 employs more complex architectures of LSTM and Multihead Attention, Model 6 with FFNN was superior in prediction. This confirms the significant improvement of BERT over word2vec and FastText. Second, all rule-based models (Models 9--11) showed better performance than the machine-learning model (Model 12). This result can be attributed to the relatively short length of the review texts and low text complexity. Alternatively, this could be due to the poor generalization performance of the machine-learning methods to imbalanced data, as restaurant ratings tend to gather at extreme values, such as 1 or 5. However, all the models scored above the chance level.

Finally, a comparison of the processing times among the LLMs showed that only Model 1, with 70B parameters, required a significantly longer processing time. Compared with Model 2, Even when analyzing only 1,000 samples, a total durations differ in more than 16 hours for a 2.4\% improvement in the prediction error. From a practical standpoint, this increase in the processing time for improvement of the accuracy cannot be considered a reasonable trade-off. It is true that Model 2 has a lower accuracy than Model 1; however, its RMSE of 0.551 indicates that the model still effectively predicts sentiment. It accurately estimates higher values as high and lower values as low, with the fastest processing speed among the proposed LLMs. 
Therefore, this study employs Llama 3 with 8B parameters and quantized in 3-bit (Model 2) for Study 2 and 3.

\begin{table*}[htb]
      \begin{center}
      \caption{Comparison of the accuracy and the processing time (Study 1, ascending in RMSE)}
      \label{tab:result1}
        \scalebox{0.88}{
\begin{threeparttable}
\begin{tabular}{r
wl{7.2 cm}wc{0.6 cm} 
rcrcrr}
\toprule
 \ccol{2}{Model Name} & Llama & \ccol{1}{\#Params}& \ccol{1}{\#Bits} & \ccol{1}{Corr.} & \ccol{1}{RMSE} & \ccol{1}{Acc.} & \ccol{1}{Time ($s$)} \\
\midrule
\multicolumn{2}{l}{\bf LLM models}\\
 & Model 01: llama-3-70b-instruct\_Q4\_K\_M & 3 & 70 billion & 4 & \cellcolor[gray]{0.8}\bf0.929 & \cellcolor[gray]{0.8}\bf0.521 & \cellcolor[gray]{0.8}\bf0.779 & $64.879 \pm 6.114$ \\
 & Model 02: llama-3-8b-instruct\_Q3\_K\_M & 3 & 8 billion & 3 & \bf0.913 & \bf0.551 & \bf0.755 & $4.560 \pm 1.768$ \\
 & Model 03: llama-3-8b-instruct\_Q4\_K\_M & 3 & 8 billion & 4 & \bf0.909 & \bf0.562 & \bf0.749 & $5.443 \pm 2.222$ \\
 & Model 04: llama-3-8b-instruct\_Q5\_K\_M & 3 & 8 billion & 5 & \bf0.892 & \bf0.617 & \bf0.756 & $5.230 \pm 2.020$ \\
 & Model 05: llama-2-7b-chat\_Q4\_K\_M & 2 & 7 billion & 4 & 0.791 & \bf0.860 & \bf0.721 & $5.986 \pm 3.850$ \\
\midrule[0.25pt]
\multicolumn{2}{l}{\bf Reference models}\\
& Model 06: BERT + FFNN& \ccol{1}{-} & \ccol{1}{-} & \ccol{1}{-} & 0.792 & 0.941 & 0.639 & \ccol{1}{-} \\
& Model 07: word2vec + LSTM + Multihead Attention & \ccol{1}{-} & \ccol{1}{-} & \ccol{1}{-} & 0.789 & 0.930 & 0.636 & \ccol{1}{-} \\
& Model 08: FastText + CNN & \ccol{1}{-} & \ccol{1}{-} & \ccol{1}{-} & 0.700 & 1.098 & 0.596 & \ccol{1}{-} \\
& Model 09: VADER & \ccol{1}{-} & \ccol{1}{-} & \ccol{1}{-} & 0.667 & 1.111 & \ccol{1}{-} & \ccol{1}{-} \\
& Model 10: TextBlob & \ccol{1}{-} & \ccol{1}{-} & \ccol{1}{-} & 0.646 & 1.120 & \ccol{1}{-} & \ccol{1}{-} \\
& Model 11: SO-CAL & \ccol{1}{-} & \ccol{1}{-} & \ccol{1}{-} & 0.661 & 1.374 & \ccol{1}{-} & \ccol{1}{-} \\
& Model 12: TF-IDF + Linear SVM & \ccol{1}{-} & \ccol{1}{-} & \ccol{1}{-} & 0.641 & 1.134 & 0.627 & \ccol{1}{-} \\
 & Model 13: Chance level & \ccol{1}{-} & \ccol{1}{-} & \ccol{1}{-} & 0.019 & 1.850 & 0.358 & \ccol{1}{-} \\
\bottomrule
\end{tabular}
\begin{tablenotes}[para,flushleft,online,normal] 
{\it Note.} Bold text indicates that the proposed model performs better than all reference models, while shading behind the indices represents the highest performance. Time represents the average duration and the standard deviation to process one review. Those not marked with an Acc. indicate cases where the prediction is in the continuous values and therefore the accuracy cannot be calculated.
\end{tablenotes}
\end{threeparttable}
        }
        \end{center}
\end{table*}

\subsection{Study 2: Integration of the Majority Voting Mechanism}
Thus far, we found the best model for sentiment analysis to assess overall ratings in Study 1. However, since evaluating the detailed aspects is significantly more difficult than evaluating the overall rating, ambiguity of the evaluation occurs during the annotation. The fact that such fluctuations were resolved by voting in the actual annotation is also an important clue for this study. Therefore, in Study 2, we incorporated majority voting by virtually generating five annotators in one LLM and examined whether the robustness of the results increased depending on the introduction of voting. As explained, the mechanism operates in two stages with $m^k$ (the presence or absence of the mention of dimension $k$), $v^k$ (the ratings of the dimension $k$), and $s^k$ (the final evaluation for dimension $k$), based on the median of the five workers (see Section 3.2 for details).

Table \ref{tab:voting} presents an example of the aggregated evaluations of the scores of five virtual workers for one review. First, as shown in the table, consistency among the workers was confirmed for most dimensions. Second, some of the evaluations were divided. For example, in $menu$, one worker rated it as 2, while the remaining four workers rated it as 3; hence, the voted sentiment was 3. In $congestion$, most of the workers evaluated that the dimension was not mentioned, whereas worker 4 evaluated it as 1. Thus, the final sentiment was set to 0, indicating that this aspect was not mentioned.

\begin{table*}[htb] 
      \begin{center}
      \caption{Results of Majority Voting (Study 2)}
      \label{tab:voting}
        \scalebox{0.94}{
\begin{threeparttable}
\begin{tabular}{
wc{0.1 cm}
wc{0.1 cm} wl{1.8 cm} 
wc{0.4 cm} wc{0.3 cm}
wc{0.3 cm} wc{0.5 cm}
wc{0.5 cm}wc{0.5 cm}
wc{1.1 cm}wc{1.1 cm}
wc{0.4 cm}wc{1.1 cm}
wc{0.6 cm}wc{0.4 cm}
wc{0.5 cm}wc{0.5 cm}
}
\toprule
& && \small rating & \small price & \small menu & \small dishes & \small dessert & \small clean & \small atmosphere & \small congestion & \small noise & \small enjoyment & \small attitude & \small child & \small couple & \small access \\
\midrule
\multicolumn{3}{l}{\bf Each Worker}\\
& \multicolumn{2}{l}{Worker 1} & 4 & 0 & 2 & 3 & 0 & 0 & 4 & 0 & 0 & 0 & 5 & 0 & 0 & 1 \\
& \multicolumn{2}{l}{Worker 2} & 4 & 0 & 3 & 3 & 0 & 0 & 4 & 0 & 0 & 4 & 5 & 2 & 3 & 1 \\
& \multicolumn{2}{l}{Worker 3} & 4 & 0 & 3 & 3 & 0 & 0 & 4 & 0 & 0 & 4 & 5 & 2 & 3 & 1 \\
& \multicolumn{2}{l}{Worker 4}  & 4 & 0 & 3 & 3 & 0 & 0 & 4 & 1 & 1 & 4 & 5 & 2 & 4 & 5 \\
& \multicolumn{2}{l}{Worker 5} & 4 & 0 & 3 & 3 & 0 & 0 & 4 & 0 & 0 & 4 & 5 & 0 & 0 & 1 \\
\midrule
\multicolumn{3}{l}{\bf Majority Voting}\\
& $m^k$&: \small $k$ is mentioned & 1 & 0 & 1 & 1 & 0 & 0 & 1 & 0 & 0 & 1 & 1 & 1 & 1 & 1 \\
& $v^k$ &: \small nonzero median & 4 & 0 & 3 & 3 & 0 & 0 & 4 & 1 & 1 & 4 & 5 & 2 & 3 & 1 \\
& $s^k$ &: \small sentiment & 4 & 0 & 3 & 3 & 0 & 0 & 4 & 0 & 0 & 4 & 5 & 2 & 3 & 1 \\
\bottomrule
\end{tabular}
\begin{tablenotes}[para,flushleft,online,normal] 
\end{tablenotes}
\end{threeparttable}
        }
        \end{center}
\end{table*}

Based on the above, Table \ref{tab:res2-2} shows a lift in the performance by incorporating the mechanism. The results indicate that that incorporating our proposed mechanism improved performance across all indices, even exceeding the average. This tendency is similar to ensemble learning which uses multiple machine-learning models to perform inference tasks and aggregates the results through statistics such as mean and median. 

\begin{table}[htb] 
      \begin{center}
      \caption{A Lift in Accuracy Depending on the Majority Voting Mechanism (Study 2)}
      \label{tab:res2-2}
        \scalebox{0.90}{
\begin{threeparttable}
\begin{tabular}{
wl{3.8 cm}wc{1.0 cm}wc{1.0 cm}wc{1.0 cm}
}
\toprule
 & \ccol{3}{Improvement (\%)}\\
\cmidrule{2-4}
Majority Voting & \ccol{1}{Corr.} & \ccol{1}{RMSE} & \ccol{1}{Acc.} \\
\midrule
 - is employed & 1.098 & 1.327 & 1.039 \\
 - in-Seed average (Seed 1--5) & 1.041 & 1.109 & 1.017 \\
 - is not-employed (baseline) & 1.000 & 1.000 & 1.000 \\
\bottomrule
\end{tabular}
\begin{tablenotes}[para,flushleft,online,normal] 
{\it Note.} Lift represents the improvement from the baseline.
\end{tablenotes}
\end{threeparttable}
        }
        \end{center}
\end{table} 

Furthermore, the difference of accuracy improvements between Study 1 and 2 is noteworthy. In Study 1, a fourteen-fold increase in the processing time resulted in improvement of 2.4\%. In contrast, in Study 2, a five-fold increase in the processing time led to that of 3.9\% using majority voting. This indicates that a medium-sized model with iterative inferences using the majority voting mechanism is significantly more efficient in terms of both training time and prediction accuracy than the larger model with a single inference.

\subsection{Study 3: Regression Analysis of the Factors Affecting the Evaluation}
Thus far, we have validated the sentiment analysis using LLMs (Study 1) and the robustness of the results using the majority voting mechanism (Study 2). In other words, we can freely extract any aspect of sentiment from review texts accurately with medium-scale LLMs using the majority voting mechanism. Finally, in Study 3, to demonstrate that further analyses are possible with the annotated data, we use regression analysis to examine how each aspect affects the overall evaluation. Since the dataset did not contain the sentiment values of each aspect, we assess the generalizability of the results from the marketing literature. For a similar analysis, a previous study \cite{paas} which examined the relationship between the emotion of the posts on the social network and the total amount of monthly donation to the university constructed linear regression models based on predicted sentiments.

We use a generalized linear model (GLM) to predict the overall rating of restaurants based on aspect-based sentiments (Table \ref{tab:dims}) as explanatory variables. In this process, we construct two GLMs with different target variables: one using the actual user rating and the other using the rating predicted by the LLM. By comparing the estimated parameters and other values of the two models, we verify that they have similar structures. Since there is a risk of multicollinearity if all the dimensions considered in this study are simultaneously used in the model, explanatory variables are selected based on Bayesian information criterion (BIC) to explore the best model.

\begin{table*}[tbh]
      \begin{center}
      \caption{A Comparison of the Regression Results (Study 3)}
      \label{tab:res3}
        \scalebox{1}{
\begin{threeparttable}
\begin{tabular}{lrlrlcrrrlcr}
\toprule
$n=1000$& \ccol{4}{$Y1$: Actual evaluation} && \ccol{4}{$Y2$: Predicted evaluation} && \ccol{1}{Diff.} \\
 \cmidrule(){2-5} \cmidrule(){7-10} \cmidrule(){12-12}
~& 
\ccol{1}{coef.} & \ccol{1}{(SE)} & \ccol{1}{$z$-value}& &~~&
\ccol{1}{coef.} & \ccol{1}{(SE)} & \ccol{1}{$z$-value} &
&~& \ccol{1}{$t$-value}\\
\midrule
intercept & 1.653 & (0.074) & 22.244 & $^\dagger$ &  & 1.477 & (0.064) & 22.797 & $^\dagger$ &  & 1.781 \\
menu & 0.144 & (0.020) & 7.154 & $^\dagger$ &  & 0.144 & (0.017) & 8.229 & $^\dagger$ &  & -0.016 \\
dishes & 0.173 & (0.019) & 9.029 & $^\dagger$ &  & 0.216 & (0.016) & 12.915 & $^\dagger$ &  & -1.682 \\
dessert & 0.054 & (0.017) & 3.205 & $^\dagger$ &  & 0.067 & (0.014) & 4.507 & $^\dagger$ &  & -0.546 \\
atmosphere & 0.087 & (0.013) & 6.629 & $^\dagger$ &  & 0.091 & (0.011) & 7.988 & $^\dagger$ &  & -0.253 \\
congestion & -0.155 & (0.027) & -5.628 & $^\dagger$ &  & -0.112 & (0.024) & -4.695 & $^\dagger$ &  & -1.155 \\
enjoyment & 0.316 & (0.017) & 18.489 & $^\dagger$ &  & 0.334 & (0.014) & 22.398 & $^\dagger$ &  & -0.785 \\
access & 0.069 & (0.024) & 2.883 & $^\dagger$ &  & 0.055 & (0.021) & 2.621 & $^\dagger$ &  & 0.450 \\
\cmidrule(r){1-1} \cmidrule(l){2-5} \cmidrule(){7-10}
psuedo-$R^2$ & \ccol{4}{~~~~~~0.768}   &  & \ccol{4}{~~~~~~0.894}   &  &  \\
AIC & \ccol{4}{2558.144}   &  & \ccol{4}{2284.276}   &  &  \\
BIC & \ccol{4}{2597.406}   &  & \ccol{4}{2323.539}   &  &  \\
\bottomrule
\end{tabular}
\begin{tablenotes}[para,flushleft,online,normal] 
{\it Note.} ~${*}: p<0.05$, ~${\dagger}: p<0.001$.
\end{tablenotes}
\end{threeparttable}
        }
\end{center}
\end{table*}

With variable selection, seven dimensions were adopted in addition to the intercept. Table \ref{tab:res3} shows the parameter estimations for the actual user ratings ($Y1$, left) and LLM-predicted ratings ($Y2$, right). First, for Y1, the results confirmed statistically significant differences at the 1\% level for all explanatory variables. In particular, the $z$ values show that $dishes$ and $enjoyment$ have a strong positive effect and $congestion$ has a strong negative effect. These are assessed as valid results, as several studies reported a similar tendency for the effects, such as "food quality" \cite{restaurant_fastfood,restaurant_chinafastfood,restaurant_quickcasual}, 
"entertainment" \cite{restaurant_boardgame}, and 
"waiting time for a meal" \cite{restaurant_quickcasual,restaurant_functionalattributes}, 
on outcomes such as perceived value, customer satisfaction, and behavioral intention. In addition to Y1, the statistical significances at the 1\% level are also confirmed for all the variables in Y2. The signs of the estimated coefficients and $z$ values are similar to those in Y1 and the $t$ tests for all explanatory variables using the coefficients and standard errors reveal no significant differences between the models (cf. Diff in Table \ref{tab:res3}). 

In summary, in Study 3, there is no significant discrepancy between the predicted and actual values of the overall evaluation, and that further analysis can be implemented using aspect-based sentiments. This indicates that the proposed model can accurately assess the impact of any dimensions on the overall evaluation. In other words, points that have not been fully evaluated in previous studies can be further analyzed by extracting more detailed perspectives from a large amount of review data. For example, in the previous study \cite{restaurant_boardgame}, the effect of the entertainment factor on the evaluation was investigated as the free provision of board games at a caf\'e. By using the proposed model, we can freely define and quantify the entertainment factor and then analyze its relationship to the overall rating.

\section{Conclusion}
In this study, we demonstrated the utility of LLMs in AbSA through a series of three investigations: 1) predicting the sentiments of multiple aspects from online review texts posted by consumers using existing pre-trained LLMs and comparing the accuracy with multiple reference models, 2) incorporating a majority voting mechanism similar to human annotation and examining its impact, and 3) constructing linear regression models between the ratings and aspects and investigating whether a discrepancy exists between the predicted and actual sentiment.

More specifically, in Study 1, by comparing the prediction accuracy and processing time of multiple pre-trained LLMs, we demonstrated that it is not necessary to use the largest-scale model in terms of the number of model parameters and the precision of quantization. Moreover, the results showed that classification is possible with higher accuracy than the well-known existing methods, such as DNNs, without any training. Thus, the proposed methodology is highly useful for practical applications because it does not require large-scale computational resources. In Study 2, we integrated the majority voting mechanism into the model from Study 1 and validated the change in the robustness. The results showed that the voting made a significant improvement on the accuracy of all the indices. When only a single model is used, LLMs sometimes fail to generate a response, which results in missing values. However, by introducing the majority voting mechanism, the missing values and prediction error are mutually filled by each model, which significantly improves the accuracy. Thus, as in previous studies \cite{annotation_gpt4}, annotation errors were reduced. In Study 3, we compared two regression models predicting the actual and assessed ratings of restaurant and showed that the results are sufficiently general, and no significant difference was confirmed between the models. 

In summary, first, in terms of the selection of the pre-trained models, multiple inferences with a medium-sized model and majority voting of the results are much faster and more accurate than a single inference on a large model with a large number of parameters. Contrary to the initial expectations, the tendency for improvement in accuracy with lower precision was confirmed, and the lower precision speeds up inference, which resulted in reducing the time burden of iterating multiple inferences. Second, using the proposed model and arbitrary dimensions, we can dynamically annotate free-form texts and facilitate the obtained structural data for advanced analysis. 
In particular, the most significant difference from the existing annotation methods is that even if the dimension used for annotation is minor or has never been used before, as long as it can be linguistically explained, we can utilize it as a dimension in the model. Thus, the proposed model has high applicability for opinion mining in questionnaires and knowledge extraction from documents across various industries, including marketing research, policymaking in administration, and patent document analysis. 

For practical implications, the proposed model provides several advantages. First, since it is constructed in a local environment, organizations can address security concerns regarding data leakage and falsification that can occur in cloud services. In particular, while companies are sometimes caught under pressure between the utilization and protection of data, this model enables them to leverage data with minimizing security risk. Second, unlike cloud services, a single purchase of a computer allows businesses to construct the model. That is, they can easily estimate the introduction and running costs, which means that companies can strategically manage their expenses for AI utilization and prevent unexpected expenditures. Third, since no additional cost occurs even if inferences are repeated with different dimensions, marketing researchers can attempt to freely explore business-useful aspects.

Finally, as the challenges of this study, while the proposed model have annotated the arbitrary dimensions, the dimensions may not necessarily provide a comprehensive understanding of the restaurant evaluation. Thus, further investigations are needed to annotate the free-form texts using the existing survey scales, such as SERVQUAL \cite{servqual} and DINESERV \cite{dineserv}.
Second, since it has been pointed out that the performance of LLMs varies depending on the instructions, it is necessary to verify the changes when the other examples are used, and the differences between zero-shot, one-shot, and a few-shots. 
Third, using the obtained structural data, various other analyses can be also implemented, such as factor analysis, correspondence analysis, clustering of users, and the development of a recommendation system. Therefore, further investigations are required to confirm whether these advanced analyses can be applied to the proposed model.

\section*{Acknowledgements}
The dataset used in this study consisted of open data for academic use. As no additional information was collected, the author did not obtain any information that could lead to the identification of individuals. The large language models used to construct the analysis model were licensed for commercial and academic use. Both the dataset and models were managed and used in an appropriate environments that comply with the terms of use of the companies from which it was made available.

This study is supported by JSPS KAKENHI (Grant Number: 24K16472).

\bibliographystyle{unsrt}
\bibliography{LLM-arxiv}

\end{document}